\documentclass[runningheads]{llncs}
\usepackage{amssymb}
\usepackage{hyperref}
\usepackage{color}

\usepackage{booktabs}
\usepackage{graphicx}
\usepackage{amsmath}
\usepackage{multirow}
\usepackage{ulem}
\usepackage[misc]{ifsym}

\hypersetup{colorlinks=true,urlcolor=blue, citecolor = blue}

\usepackage{subcaption}

\begin{document}

\title{
PECon: Contrastive Pretraining to Enhance Feature Alignment between CT and EHR Data for Improved Pulmonary Embolism Diagnosis
}
\titlerunning{PECon}
% \author{*}
% \institute{*}

\author{Santosh Sanjeev\textsuperscript\Letter \inst{1}\orcidID{0000-0003-3664-3844} \and
Salwa K. Al Khatib\inst{1}\orcidID{0000-0002-9588-8473}\ \and
Mai A. Shaaban\inst{1}\orcidID{0000-0003-1454-6090} \and
Ibrahim Almakky\inst{1}\orcidID{0009-0008-8802-7107} \and
Vijay Ram Papineni\inst{2}\orcidID{0000-0002-6162-3290} \and
Mohammad Yaqub\inst{1}\orcidID{0000-0001-6896-1105}}

% \orcidID{0000-0003-3664-3844}
% \orcidID{1111-2222-3333-4444}
% \orcidID{0000-0003-1454-6090}
% \orcidID{2222--3333-4444-5555}
% \orcidID{2222--3333-4444-5555}
% \orcidID{0000-0001-6896-1105}}
%
\authorrunning{Sanjeev et al.,}
\institute{Mohamed bin Zayed University of Artificial Intelligence, Abu Dhabi, UAE \and Sheikh Shakbout Medical City, Abu Dhabi, UAE \\
\email{\{santosh.sanjeev,salwa.khatib,mai.kassem,ibrahim.almakky, \\mohammad.yaqub\}@mbzuai.ac.ae, vpapineni@ssmc.ae}}
\maketitle              % typeset the header of the contribution
\begin{abstract}
Previous deep learning efforts have focused on improving the performance of Pulmonary Embolism (PE) diagnosis from Computed Tomography (CT) scans using Convolutional Neural Networks (CNN). However, the features from CT scans alone are not always sufficient for the diagnosis of PE. CT scans along with electronic heath records (EHR) can provide a better insight into the patient’s condition and can lead to more accurate PE diagnosis. In this paper, we propose \textbf{P}ulmonary \textbf{E}mbolism Detection using \textbf{Con}trastive Learning (\textbf{PECon}), a supervised contrastive pretraining strategy that employs both the patient’s CT scans as well as the EHR data, aiming to enhance the alignment of feature representations between the two modalities and leverage information to improve the PE diagnosis. In order to achieve this, we make use of the class labels and pull the sample features of the same class together, while pushing away those of the other class. Results show that the proposed work outperforms the existing techniques and achieves state-of-the-art performance on the RadFusion dataset with an F1-score of 0.913, accuracy of 0.90 and an AUROC of 0.943. Furthermore, we also explore the explainability of our approach in comparison to other methods. Our code is publicly available at \url{https://github.com/BioMedIA-MBZUAI/PECon}.
%Furthermore, we also evaluate the performance of our imaging-only model on another private dataset to understand the impact and generalizability of our pretraining strategy.

% Pulmonary Embolism (PE) is the third most common cause of cardiovascular death after coronary artery disease and stroke. With the rise of deep learning, research works have tried to improve the performance of PE diagnosis from Computed Tomography (CT) scans using Convolutional Neural Networks (CNN). Unfortunately, the information from the CT scans alone is inadequate in the diagnosis of PE. The CT scans along with the patient’s electronic heath record (EHR) data can provide a much deeper insight into the patient’s condition and can help in the early diagnosis of PE. In this paper, we propose a self-supervised contrastive pretraining strategy that uses both the patient’s CT study as well as the EHR data, aiming to enhance the cohesion between the different modality feature representations and leverage information to improve the PE diagnosis. For our work we have considered the RadFusion dataset which includes around 1837 studies from 1794 patients. Results show that the proposed work outperforms the existing techniques and achieves state-of-the-art performance on the Radfusion dataset with an F1-score of 91.6\% and an accuracy of 90\%. Furthermore, we also evaluate the performance of our imaging-only model on another private dataset to understand the impact and generalizability of our pretraining strategy.

\end{abstract}

\section{Introduction}
\label{sec:intro}
Pulmonary Embolism (PE) is an acute cardiovascular disorder considered the third most common cause of cardiovascular death after coronary artery disease and stroke. Despite advances in diagnosis and treatment over the past $30$ years, PE has high early mortality rates \cite{Blohlvek2013PulmonaryEP}, with nearly $100k$ to $200k$ deaths in the US each year \cite{TarboxAbigail7386237}. Unfortunately, individuals diagnosed with PE frequently encounter a long delay before receiving a diagnosis, and approximately $25\%$ of patients are initially misdiagnosed \cite{Hendriksen2017,Alonso2010}. Hence, it is crucial to provide radiologists and clinicians with tools that can help them with the diagnosis.
% Diagnosis of PE typically involves imaging tests such as CT scans, chest x-rays, and ultrasound scans, as well as blood tests to measure markers of clotting and inflammation. 

%It affects the right ventricular of the heart, thereby proving to be a life-threatening condition.  Due to this, patients die within a few hours, which emphasizes the importance of early diagnosis of PE \cite{Blohlvek2013PulmonaryEP}.

% Pulmonary embolism (PE) is a potentially life-threatening condition that occurs when a blood clot (thrombus) forms in one of the deep veins in the body, usually in the legs, and then travels to the lungs \cite{PE}. The clot can block blood flow in the pulmonary arteries, leading to impaired oxygen exchange and potentially causing damage to the lungs and other organs. Diagnosis of PE typically involves imaging tests such as CT scans, chest x-rays, and ultrasounds, as well as blood tests to measure markers of clotting and inflammation.

% PE is considered the third most common cause of cardiovascular death after coronary artery disease and stroke. It is an acute cardiovascular disorder having high early mortality rates despite the advances in diagnosis and treatment over the past 30 years \cite{Blohlvek2013PulmonaryEP}. PE is responsible for nearly $100,000$ to $200,000$ deaths in the US each year \cite{TarboxAbigail7386237}. It affects the right ventricular of the heart, thereby proving to be a life-threatening condition. Due to this, patients die within a few hours, which emphasizes the importance of early diagnosis of PE \cite{Blohlvek2013PulmonaryEP}.

Prior to deep learning, many efforts focused on using traditional feature extraction methods for PE diagnosis from Computed Tomography (CT) scans \cite{Masutani2002,liang2007}. However, more recently research efforts have investigated improving the performance of PE diagnosis using deep Convolutional Neural Networks (CNNs) \cite{lin2019,Khachnaoui2022} and attention mechanisms \cite{suman2021}. More specifically, PENet \cite{penet} was introduced as a 3D CNN designed to detect PE using multiple CT slices. Such use of 3D convolutions allows the network to consider information from multiple slices when making predictions, which is important for diagnosing PE, as its presence is not limited to a single CT slice.

Features from CT scans alone could be insufficient for PE diagnosis. Therefore, CT scans along with the patient's Electronic Health Records (EHR) data can provide a better insight into the patient's condition and can help improve the diagnosis of PE. Previous studies have shown improved performance when combining demographic and clinical data with medical imaging data for various medical conditions such as Alzheimer's disease and skin cancer \cite{Li2019,8333693}. However, multimodal fusion is a non-trivial task, which can have varying results based on the fusion approach. Therefore, Huang et al. \cite{huang2020multimodal} compared  different multimodal fusion approaches that combine inputs from both CT scans and EHR data to diagnose the presence of PE and explore optimal data selection and fusion strategies. Their results show that multimodal end-to-end deep learning models combining imaging and EHR provide better discrimination of abnormalities than using either modality independently. To this effect, Zhou et al. \cite{radfusion} released RadFusion, a multimodal dataset containing both EHR data and high-resolution CT scans labeled for PE. They assess the fairness properties across different subgroups and results suggest that integrating EHR data with medical images can improve classification performance and robustness without introducing large disparities between population groups.

% The work in \cite{huang2020multimodal} involved comparing several architectures of multimodal fusion models that can use both pixel data from volumetric CT Pulmonary Angiography scans and clinical patient data from electronic medical records to classify cases of PE automatically. 
% The main contributions in this work include developing and evaluating various end-to-end deep learning models for detecting PE using both CT imaging and EMR data. 
% % The best-performing model is a late fusion model using 3D CNN and ElasticNet, which achieved an AUROC score of $0.962$. 

Contrastive learning has been previously employed in medical imaging. Recently, Zhang et al. \cite{contrastivemedical} made use of available pairs of images and text usually in medical contexts, and proposed Contrastive VIsual Representation Learning from Text (ConVIRT) \cite{contrastivemedical}. The approach involves pretraining medical image encoders with paired text via a bidirectional contrastive objective. Nevertheless, their text encoder employs a general-purpose lexicon, resulting in issues with unknown words when processing a medical text. Although the division of terms into word pieces helps alleviate this problem, it results in the fragmentation of common biomedical terminology, leading to suboptimal results \cite{zhang2023large,gu2021domain}. Furthermore, \cite{supcon} has shown that constrastive learning between two images (same modality) in a fully-supervised setting can outperform the self-supervised contrastive pretraining by leveraging the label information. In this paper, we propose \textbf{P}ulmonary \textbf{E}mbolism Detection using \textbf{Con}trastive Learning (\textbf{PECon}), a novel fully supervised contrastive pretraining framework between 3D CT scans and EHR data. PECon uses a supervised contrastive learning objective to train a neural network to encode CT scans and EHR into a joint embedding space. In this space, we establish cross-modal alignment, through a fully supervised contrastive (SC) pretraining aiming to maximise the feature alignment between CT and EHR data.
% where the similarity between the two modalities is maximized.
To the best of our knowledge we are the first to introduce a SC pretraining using different modalities (EHR,CT) in medical domain to maximize alignment between the feature representations. Unlike \cite{supcon}, which uses augmentations to get 2 views of the same image(single modality), we use CT and EHR as 2 views(multimodal) of the same patient.
We demonstrate the effectiveness of our approach on a real-world dataset of PE patients, where we show that our pretrained model significantly outperforms state-of-the-art baselines. 
% To this end, we propose \textbf{P}ulmonary \textbf{E}mbolism Detection using \textbf{Con}trastive Learning (\textbf{PECon}), a supervised contrastive pretraining method to maximize feature alignment between the CT scans and the better-performing EHR data. Formally, 
% \sout{In this space, the aim is to maximize feature alignment between CT scans and EHR data.} 
The main contributions of this work can be summarized as follows: 
\begin{itemize}
    \item We propose a supervised contrastive pretraining method that better aligns the image and EHR embeddings thereby achieving enhanced alignment between the features of both modalities. 
    \item We demonstrate how the proposed method enhances the performance of multimodal PE detection and achieves state-of-the-art results on the RadFusion \cite{radfusion} dataset.
    % \item We improve the performance of PENet \cite{penet} on a private dataset, which shows that the proposed pretraining method improves the generalizability of this imaging-only model.
    % \item We assess model generalizability by evaluating its performance on a private dataset that includes CT scans from a diverse set of ethnicities.
\end{itemize}

\begin{figure}[t]
\centering
\includegraphics[width=0.95\textwidth]{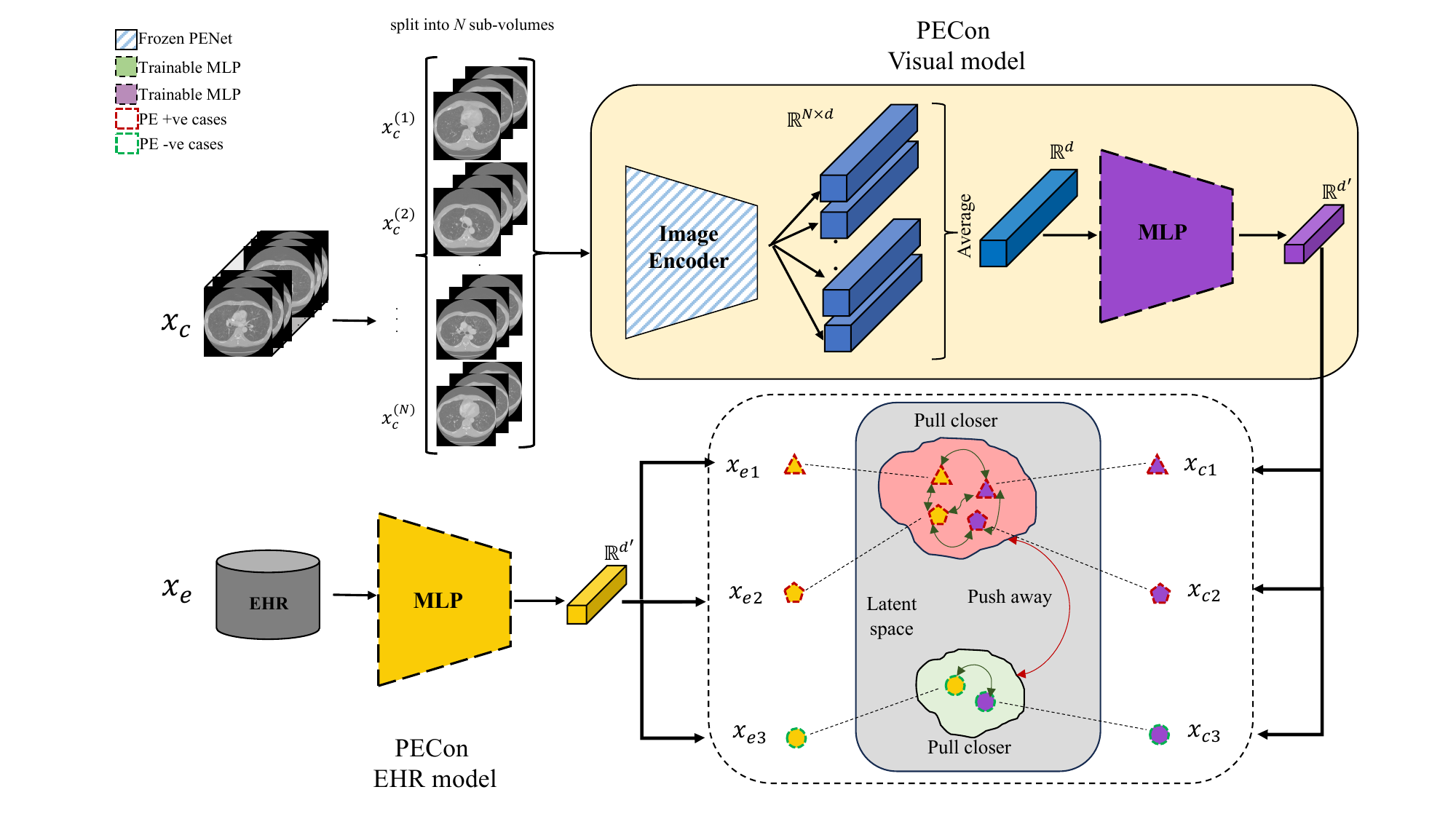}
\caption{PECon: a self-supervised contrastive pretraining method for improved PE diagnosis. The framework consists of a visual branch with a frozen image encoder backbone and a trainable MLP (top part) and an EHR branch with a trainable MLP (bottom left part). The output representations of both MLPs are later used for supervised contrastive loss (bottom right part). $(x_{c1}, x_{e1})$, $(x_{c2}, x_{e2})$, $(x_{c3}, x_{e3})$ are the CT and EHR embeddings of 3 patients in a given batch.}
\label{fig:architecture}
\end{figure}

\section{Methodology}
\label{sec:methodology}

In this section, we discuss the details of PECon, the proposed pretraining method illustrated in Fig.~\ref{fig:architecture}, and the fine-tuning stage that follows. The pretraining stage involves a supervised contrastive learning step, where, for each anchor feature extracted from a CT scan volume or an EHR record, all other features having the same label as this feature in the batch are \textit{pulled} to it, and all those having the opposite label are \textit{pushed} away. As such, given a pair of a CT scan and corresponding EHR data $(x_{c}, x_{e})$ belonging to the same patient, our goal is to learn a function $M(x_{c}, x_{e}) = \hat{y}_{x_{c}, x_{e}}$ that uses this paired input to approximate $y$, the ground truth PE diagnosis of patient. We set a pre-text task whose objective is to learn an image projection head $f_{c}$ and an EHR projection head $f_{e}$ that respectively map $x_{c}$ and $x_{e}$ to $d^{'}$-dimensional vectors $z$. More precisely, projection head $f_{c}$ takes as input a $d$-dimensional embedding, which is the result of averaging $N \times d$-dimensional embeddings from an image encoder. The image encoder is used to generate a $d$-dimensional embedding for each of the $N$ sub-volumes belonging to a single CT scan ($x_{c}^{(1)}$, $x_{c}^{(2)}$, .. ,$x_{c}^{(N)}$ as shown in Fig. \ref{fig:architecture}). For a given input batch of data, \cite{supcon} applies data augmentation twice and obtains two copies of the batch. Alternatively, in PECon, considering a batch containing $B$ pairs of $(x_{c}, x_{e})$, the training objective of PECon is a supervised contrastive loss with the following expression:

\begin{equation}
    \label{eq:supcon}
\mathcal{L}_{PECon}= \sum_{i=1}^{2B} \frac{-1}{|P(i)|} \sum_{p\in P(i)} log\frac{exp(\langle z_{i}.z_{p}\rangle / \tau)}{\sum_{a\in A(i)} exp(\langle z_{i}.z_{a} \rangle / \tau)}
\end{equation}

% \begin{equation}
%     \label{eq:supcon}
% \mathcal{L}_{PECon}= \mathcal{L}_{PECon}^{c} + \mathcal{L}_{PECon}^{e}
% \end{equation}
% The individual loss terms $\mathcal{L}_{PECon}^{c}$ and $\mathcal{L}_{PECon}^{e}$ can be expressed as:

% \begin{equation}
%     \label{eq:supcon}
% \mathcal{L}_{PECon}^{c}= \sum_{i=1}^{B} \frac{-1}{|P(c_{i})|} \sum_{p\in P(c_{i})} log\frac{exp(\langle c_{i}.p\rangle / \tau)}{\sum_{a\in A(c_{i})} exp(\langle c_{i}.a \rangle / \tau)}
% \end{equation}

% \begin{equation}
%     \label{eq:supcon}
% \mathcal{L}_{PECon}^{e}=  \sum_{i=1}^{B}\frac{-1}{|P(e_{i})|}\sum_{p\in P(e_{i})} log\frac{exp(\langle e_{i}.p\rangle / \tau)}{\sum_{a\in A(e_{i})} exp(\langle e_{i}.a \rangle / \tau)}
% \end{equation}

% \begin{equation}
%     \label{eq:supcon}
% \mathcal{L}_{PECon}^{c}= \sum_{i=1}^{B} \frac{-1}{|P(i)|} \sum_{p\in P(i)} log\frac{exp(\langle c_{i}.p\rangle / \tau)}{\sum_{a\in A(i)} exp(\langle c_{i}.a \rangle / \tau)}
% \end{equation}

% \begin{equation}
%     \label{eq:supcon}
% \mathcal{L}_{PECon}^{e}=  \sum_{j=B}^{2B}\frac{-1}{|P(j)|}\sum_{p\in P(j)} log\frac{exp(\langle e_{j}.p\rangle / \tau)}{\sum_{a\in A(j)} exp(\langle e_{j}.a \rangle / \tau)}
% \end{equation}

%$i \in I \equiv\{1...D\} $ is the index of an arbitrary feature pair corresponding to an input pair $(x_{c}, x_{e})$ within a given batch used as the anchor features
where $A(i)$ is the set of all $z$ features extracted from $x_{c}$ and $x_{e}$ samples in the batch distinct from feature $i$, $P(i) \equiv \{p \in A(i): y_{p}=y_{i} \}$ is the set of features within the batch having the same ground truth label as feature $i$, $|.|$ is the set cardinality, and $\tau \in \mathcal{R}^{+}$ is a temperature parameter.

The visual  projection head $f_{c}$ and the EHR  projection head $f_{e}$ are then separately fine-tuned for classification by minimizing the cross-entropy loss. For this, the multimodal prediction is given by: 
% For final evaluation, we use late fusion of both models as follows:
\begin{equation}
    \label{eq:final-proba}
    \hat{y}_{x_{c},x_{e}}= \lambda \hat{y}_{x_{c}}+ (1-\lambda)\hat{y}_{x_{e}}
\end{equation}
where $\lambda$ is a modality weighting hyper-parameter $\in [0,1]$, which can be used to set the influence each modality model will have on the overall multimodal classification.

\section{Experimental Setup}
\subsection{Dataset}
% In this work, we use two datasets, the first being the publicly available RadFusion dataset \cite{radfusion} and the other is privately collected dataset. 
In this work, we use the RadFusion dataset \cite{radfusion}, which consists of both CT scans and EHR data of the patients with and without PE. It consists of $1,837$ axial CTPA exams from $1794$ patients captured with $1.25mm$ spacing. The number of negative PE cases is $1,111$, while the number of positive PE cases is $726$. The EHR data consists of demographic features, vitals such as systolic and diastolic blood pressure, body temperature, $641$ unique inpatient and outpatient medications, $141$ unique diagnosis groupings, and laboratory tests of around $22$ categories. This information is represented with a binary presence/absence as well as the latest value of the test.  EHR in the dataset is structured (categorical, numerical) without text. PE cases are mainly of 3 types: central, segmental and subsegmental based on the location of PE within the arterial branches. The number of cases of central, segmental and subsegmental PE are 257, 387 and 52 respectively.  For a fair comparison, we followed the standard split of Radfusion\cite{radfusion} which was, training, validation, and testing splits of $80\%$, $10\%$, $10\%$ respectively. The scans are normalised between $[-1, 1]$ and the EHR data is normalised between $[0,1]$.

\subsection{Implementation Details}
\label{subsec:imp_details}
\textbf{Pretraining stage.}
Pretraining is carried out in an end-to-end fashion with an SGD optimizer, using an initial learning rate (LR) of $0.1$, $\tau = 0.8$, a batch size of $128$, and for $100$ epochs on a single NVIDIA RTX A6000 GPU. We select the weights of the epoch with the lowest validation loss for later fine-tuning. As for data augmentation, the 3D scans are randomly (1) cropped along their widths and heights to $192\times 192$, (2) jittered up to 8 slices along the depth axis, and (3) rotated up to $15^{\circ}$, as is done in \cite{penet}. The image encoder architecture we adopt is PeNet \cite{penet}. The PENet we use is an end to end 3D CNN model consisting of 4 encoder blocks primarily constructed from 3D convolutions with skip connections and squeeze-and-excitation blocks. The PENet backbone is initialized with weights trained on RadFusion, pretrained on Kinetics dataset \cite{kinetics} and it remains frozen during the pretraining. Input to the PENet models are subvolumes consisting of $24$ slices, where PENet outputs a $2048$ vector for each subvolume. The feature vector input to the visual projection head is an average of the $2048$ embeddings generated for each subvolume. The image projection head is an MLP consisting of $2$ hidden layers with $512$ and $256$ neurons respectively outputting a feature vector of size $128$. Similarly, the EHR projection head consists of $1$ hidden layer with $128$ neurons. Image size and number of slices were chosen based on the PENet\cite{penet}, which provided the best results and allowed for a fair comparison. The modality weighting parameter was selected empirically by testing with different values and choosing the best performing on validation set. We explored different embedding sizes (128,256,512) with performance stagnation after 128. Similarly other hyper-parameters have been experimentally selected. 
\textbf{Fine-tuning stage.}
In the fine-tuning stage, the visual projection head and the EHR projection head  are separately fine-tuned on the RadFusion dataset. The imaging model is fine-tuned for $25$ epochs with an SGD optimizer, using a LR of $0.01$ and a stepLR scheduler with a step size of 20. The EHR model is fine-tuned for $25$ epochs using SGD optimizer with a LR of $0.1$ and a stepLR scheduler with a step size of 10. We report results for a $\lambda$ of $0.375$ and the ablation is shown in the Supplementary.

\section{Results and Discussion}
\label{sec:exps}
% \subsection{Comparison to the state-of-the-art}

\begin{table}[h]
\centering
\caption{Performance Comparison of the imaging-only, EHR-only and multimodal fusion models with the state-of-the-art}
\label{table1:results}
\begin{tabular}{l|c|lccc} 
\toprule
Modality                                                                            & \begin{tabular}[c]{@{}c@{}}Include\\ subsegmental\end{tabular} & \multicolumn{1}{c}{Models} & \multicolumn{1}{c}{Accuracy} & \multicolumn{1}{c}{F1 score} & \multicolumn{1}{c}{AUROC}  \\ 
\hline
\multirow{2}{*}{CT}                                                 & \multirow{6}{*}{$\checkmark$}                                  & PENet \cite{penet}                     & 0.689                        & 0.677                        & \textbf{0.796}                      \\
                                                                                    &                                                                & \textbf{PECon Visual model}       & \textbf{0.726}               & \textbf{0.752}              & 0.775                      \\
\multirow{2}{*}{EHR}                                                     &                                                                & ElasticNet \cite{radfusion}                 & 0.837                        & 0.850                        & \textbf{0.922}             \\
                                                                                    &                                                                & \textbf{PECon EHR model}       & \textbf{0.858}               & \textbf{0.872}               & \textbf{0.922}             \\
\multirow{2}{*}{\begin{tabular}[c]{@{}c@{}}Multimodal \end{tabular}} &                                                                & Radfusion \cite{radfusion}                  & 0.890                         & 0.902                        & \textbf{0.946}                      \\
                                                                                    &                                                                & \textbf{PECon Multimodal}       & \textbf{0.900}                 & \textbf{0.913}               & 0.943                      \\ 
\hline
\multirow{2}{*}{CT}                                                 & \multirow{6}{*}{$\times$}                                      & PENet \cite{penet}                        & \textbf{0.759}                        & 0.735                        & \textbf{0.842}                      \\
                                                                                    &                                                                & \textbf{PECon Visual model}       & \textbf{0.759}              & \textbf{0.766}               & 0.817                      \\
\multirow{2}{*}{EHR}                                                     &                                                                & ElasticNet \cite{radfusion}                & 0.877                        & 0.877                        & \textbf{0.932}                      \\
                                                                                    &                                                                & \textbf{PECon EHR model}       & \textbf{0.883}               & \textbf{0.883}               & 0.930                      \\
\multirow{2}{*}{\begin{tabular}[c]{@{}c@{}}Multimodal \end{tabular}} &                                                                & Radfusion \cite{radfusion}                 & 0.895                        & 0.895                        & \textbf{0.962}                      \\
                                                                                    &                                                                & \textbf{PECon Multimodal}       & \textbf{0.914}               & \textbf{0.918}               & 0.961                      \\
\bottomrule
\end{tabular}
\end{table}

\begin{figure}[!h]
\centering
\includegraphics[width=0.85\textwidth]{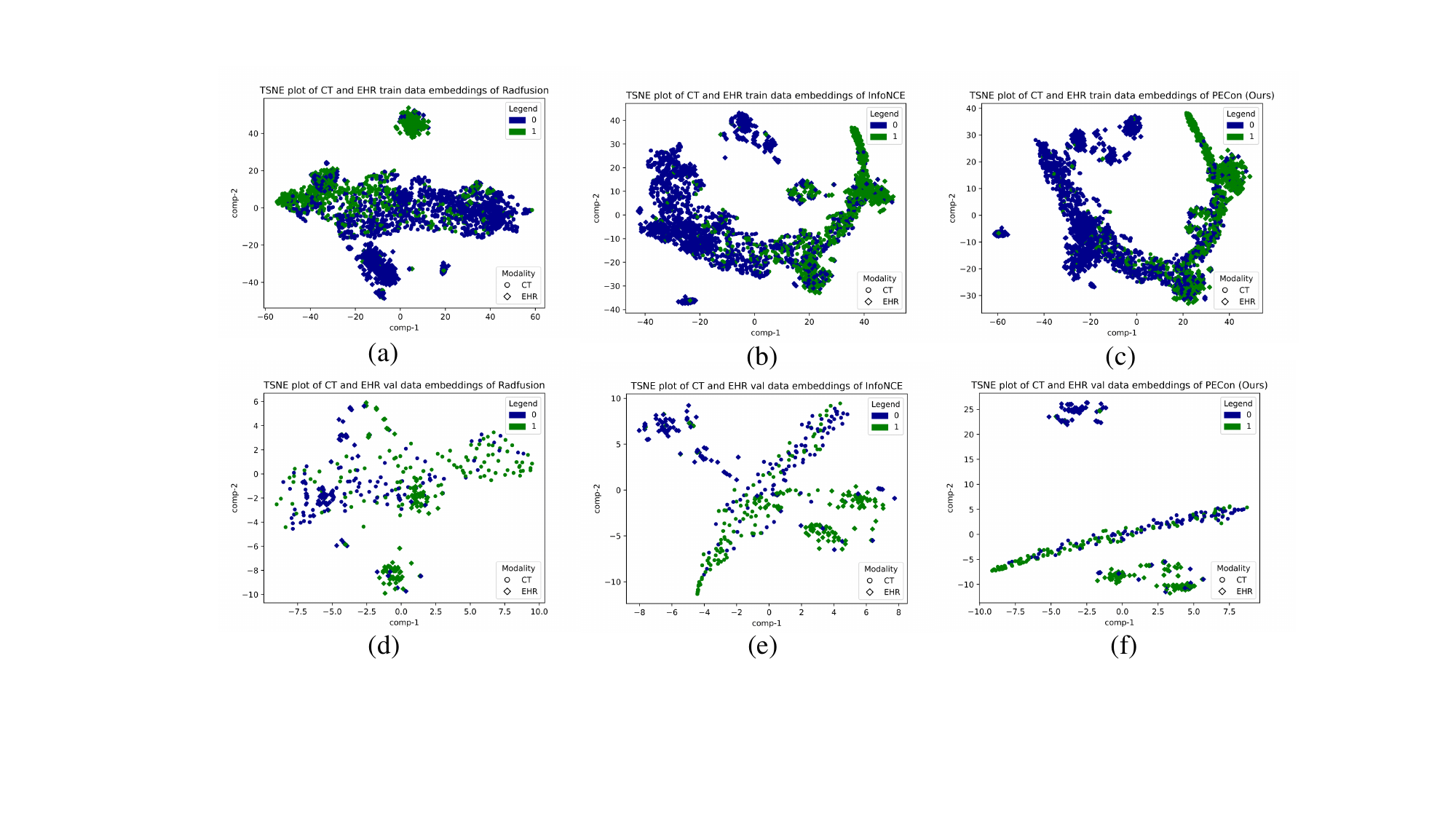}
\caption{Illustration of TSNE plots. (a) Train set embeddings of Radfusion, (b) Train set embeddings of InfoNCE, (c) Train set embeddings of PECon (Ours), (d) Val set embeddings of Radfusion, (e) Val set embeddings of InfoNCE and (f) Val set of embeddings of PECon} 
\label{fig:tsne}
\end{figure}

We consistently use the RadFusion test set to evaluate our approach and compare it with existing approaches. Table \ref{table1:results} compares the performance of our approach with the benchmark models. We observe that PECon achieves state-of-the-art performance on CT, EHR, as well as the multimodal setting. The boost in multimodal performance demonstrates the effectiveness of the approach in improving class separability in the latent space. Further, the improvement on the individual modalities shows a good separation even at the modality level, where samples belonging to the same class are pulled together and samples belonging to different classes are pushed apart. Although, we observe a small dip in the AUROC score in the visual and multimodal settings, our focus is on the F1-score as it provides a better measure of performance considering the slight class imbalance. 
% combined EHR and CT representation 
Due to the questionable clinical value of subsegmental PE \cite{Albrecht2016Nov}, we evaluate and compare the results with the benchmark models including and excluding subsegmental PE from the test set. We observe similar performance gains when including or excluding the subsegmental cases from the test set. However, excluding subsegmental cases from the training data for PECon during both pretraining and fine-tuning stages results in the best performance. We test the effect of excluding the sub-segmental cases in either the pretraining or the fine-tuning stages, while still including them in the evaluation. Results in Table~\ref{table:ablation-1} show that excluding samples from both stages achieves the best results.
% using Supervised Contrastive Loss.
% To evaluate the generalizability of our proposed approach we test the performance of the fine-tuned PECon imaging-only model on a private dataset. The results are shown in Table \ref{results:private_dataset}, thereby justifying the proposed self-supervised pretraining strategy.

% \usepackage{multirow}
% \begin{table}[t]
% \centering

% \caption{Ablation studies on the inclusion of subsegmental cases in the pre-training and finetuning stages.}
% \label{table:ablation-1}
% \begin{tabular}{ccccc}
% \toprule
% Pre-training & Finetuning & Accuracy & F1 score & AUROC              \\ \hline
%  $\checkmark$      & $\checkmark$      &  0.868              &   0.886             &   0.928            \\
%  $\checkmark$      & $\times$          &   0.858             &    0.878       &     0.920          \\
% $\times$          &  $\checkmark$      &    0.868            &     0.885          &     0.923          \\
% \textbf{$\times$} & \textbf{$\times$} &  \textbf{0.895} & \textbf{0.905} &  \textbf{0.940} \\ \bottomrule
% \end{tabular}
% \end{table}

\begin{table}[t]
\centering

\caption{Ablation studies on the inclusion of subsegmental cases in the pre-training and finetuning stages.}
\label{table:ablation-1}
\begin{tabular}{ccccc}
\toprule
Pre-training & Finetuning & Accuracy & F1 score & AUROC              \\ \hline
 $\checkmark$      & $\checkmark$      &  0.889              &   0.905            &   0.931            \\
 $\checkmark$      & $\times$          &   0.878             &    0.893       &     0.941          \\
$\times$          &  $\checkmark$      &    0.873            &     0.887          &     0.920          \\
\textbf{$\times$} & \textbf{$\times$} &  \textbf{0.900} & \textbf{0.913} &  \textbf{0.943} \\ \bottomrule
\end{tabular}
\end{table}
\begin{figure}[t]
\centering
\includegraphics[width=0.5\textwidth]{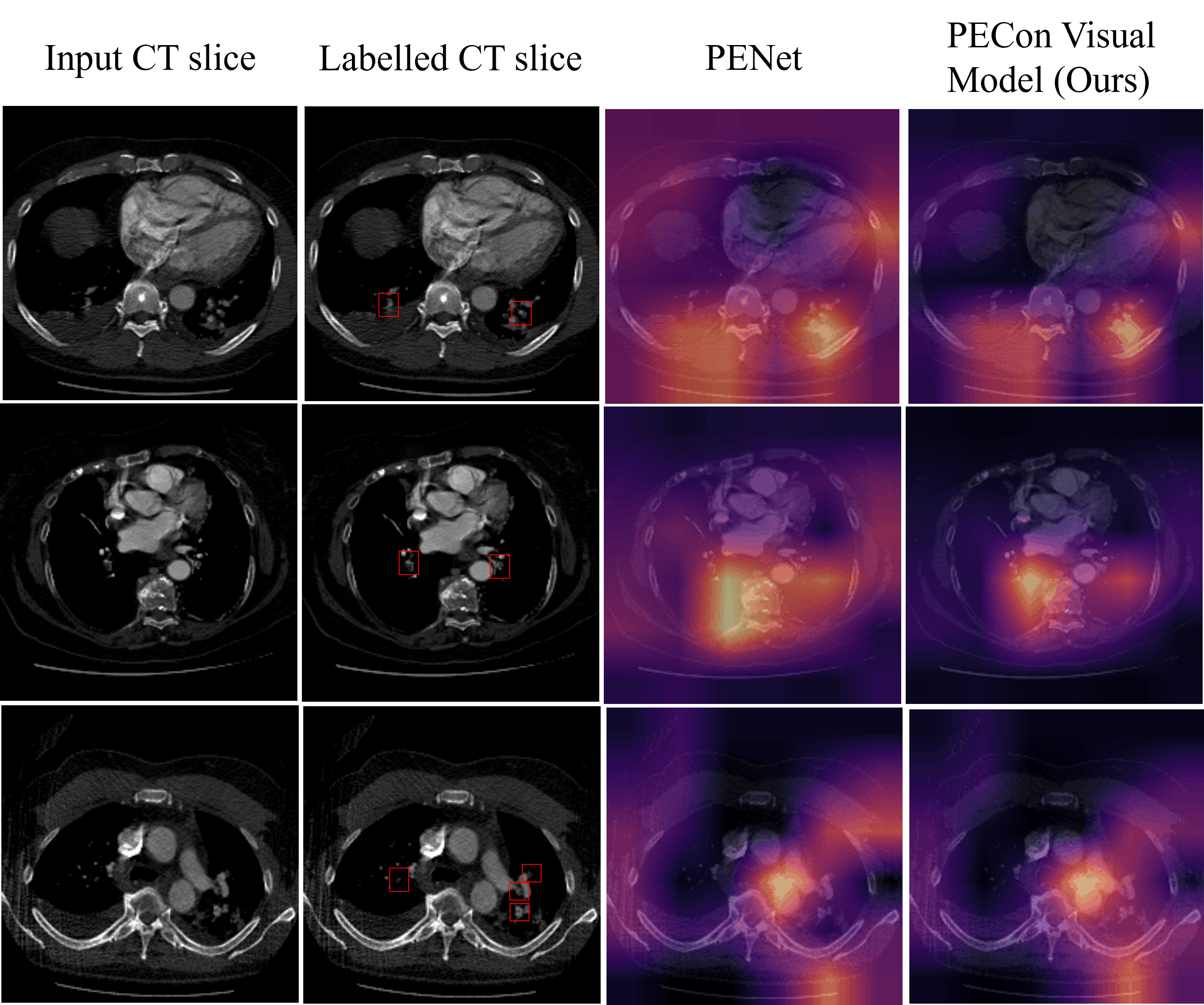}
\caption{Class activation map representations of PENet and PECon on sample CT slices from the RadFusion dataset.}
\label{fig:gradcams}
\end{figure}

% \begin{table}[]
% \centering
%     \caption{\centering Performance Comparison of imaging-only models on a Private dataset}
%     \begin{tabular}{l|l|l|l}
%     \toprule
%     Models & Accuracy & F1 score & AUROC \\ \hline
%     PENet & 0.5 & 0.5 & 0.5 \\
%     \textbf{PECon(Ours)} & \textbf{0.5} & \textbf{0.5} & 0.5 \\ \bottomrule
%     \end{tabular}
%     \label{results:private_dataset}
% \end{table}

% In Figure~\ref{fig:tsne}, we visualize and compare the TSNE plots of Radfusion and PECon (Ours). We also visualize results of PECon with a self-supervised contrastive learning approach InfoNCE \cite{infonce}. We observe that during training, the CT and EHR positive and negative features respectively are being pulled together, thereby making the data more separable and forming clusters, whereas the Radfusion train and test data do not have a good separation. To enhance the explainability of our approach, we visualize class activation maps of three CT slices pertaining to positive scans from the RadFusion dataset using PENet and PECon in Figure~\ref{fig:gradcams}. PECon consistently demonstrates better and more precise localization of the PE.
In Figure~\ref{fig:tsne}, we visualize and compare the TSNE plots of Radfusion and PECon (Ours) for the train and validation sets. We also visualize results of PECon with InfoNCE \cite{infonce}. We observe that pretraining using a contrastive loss enhances the feature alignment between CT scan and EHR features, thereby making the data more separable and forming clusters. On the other hand, the RadFusion train and validation do not have a good separation. Furthermore, the use of supervision in PECon pulled together sample features of the same class while pushing away the features of different classes. This is lacking in the case of InfoNCE since it is a fully self-supervised method that is agnostic to the class labels of the features. To enhance the explainability of our approach, we use Grad-CAM \cite{gradcam} to visualize class activation maps of three CT slices pertaining to positive scans from the RadFusion dataset using PENet and PECon in Figure~\ref{fig:gradcams}. In the case of PENet, some attention is put on irrelevant regions around the lungs, whereas PECon imaging model consistently demonstrates better and more precise localization of the PE. This shows that combining the two modalities using our approach enables the visual model to learn more precise features from the input volumes. We further compare our pretraining stage with other contrastive learning methods. Results in Table~\ref{table:ablation-2} show that PECon achieves better performance. This demonstrates that the inclusion of supervision with contrastive learning is more effective for PE classification compared to self-supervised pretraining methods. The hyper-parameters for each method are reported in the Supplementary.

\label{subsec:ablation}

\begin{table}[t]
\centering
\caption{Ablation studies on the contrastive learning strategy}
\label{table:ablation-2}
\begin{tabular}{lccc}
\hline
\begin{tabular}[c]{@{}l@{}}Contrastive Learning\\ strategy\end{tabular} & Accuracy     & F1 score       & AUROC          \\ \hline
ConViRT (InfoNCE) \cite{contrastivemedical}                                                                     & 0.892\textpm0.003        & 0.907\textpm0.002          & 0.938\textpm0.001          \\
Barlow Twins \cite{zbontar2021}                                                             & 0.889\textpm0.005        & 0.904\textpm0.003          & 0.930\textpm0.004          \\
PECon                                                              & \textbf{0.898\textpm0.003} & \textbf{0.912\textpm0.002} & \textbf{0.939\textpm0.003} \\ \hline
\end{tabular}
\end{table}

\section{Conclusion and Future Work}
\label{sec:conc}
In this work, we present (\textbf{PECon}), a supervised contrastive pretraining framework for multimodal prediction of PE to enhance the alignment between the CT and EHR latent representations. To accomplish this, we start from an anchor feature extracted from CT scans and EHR data and pull features from the same class to it and push away those from the other class. We trained and validated our approach on the RadFusion dataset, and report state-of-the-art results. Additionally, we show that our model attends to more precise and localized regions in the CT scans through class activation maps.

The limitations of this work include testing on other datasets to assess the generalizability of our approach. Furthermore, our study was limited to the impact of the contrastive pretraining loss on the late fusion multimodal approach, where in the future it would be important to study this impact on early and joint fusion approaches. Further, attention techniques will be explored for feature aggregation.

% We acknowledge that medical text modality is an important source of information that can aid in accurate diagnosis. Future work should incorporate medical text modality to improve the performance of our model. Furthermore, we have tested our model on one large-scale dataset, and although the results are promising, further testing on more diverse datasets is necessary to verify the generalizability of our model.

%
% the environments 'definition', 'lemma', 'proposition', 'corollary',
% 'remark', and 'example' are defined in the LLNCS documentclass as well.
%

\bibliographystyle{splncs05}
\bibliography{refs}




% LNCS Homepage, \url{http://www.springer.com/lncs}. Last accessed 4
% \end{thebibliography}
\end{document}

\title{PECon: Contrastive Pretraining to Enhance Feature Alignment between CT and EHR Data for Improved Pulmonary Embolism Diagnosis}

\author{}
\institute{}
\maketitle 

\section{Supplementary Material}

\begin{table}[]
\caption{The hyperparameters used to train the models using the different contrastive learning approaches. The Weighting factor in InfoNCE loss during pretraining contributes to to the weight given to the loss from CT to EHR. In Barlow Twins experiment, it weighs the off diagonal elements. For both the experiments in finetuning, the weighting contributes to the CT model. }
\centering
\begin{tabular}{lccccc}
\hline
Experiment                    & Stages      & Epochs & Optimizer            & Learning Rate                                            & Weighting factor \\ \hline
\multirow{2}{*}{ConViRT (InfoNCE)}      & Pretraining & 100    & \multirow{2}{*}{SGD} & 0.1                                                      & 0.5              \\
                              & Finetuning  & 25     &                      & 0.01(CT), 0.1(EHR) & 0.495            \\ \hline
\multirow{2}{*}{Barlow Twins} & Pretraining & 300    & \multirow{2}{*}{SGD} & 0.1                                                      & 0.0051           \\
                              & Finetuning  & 25     &                      & 0.01(CT), 0.1(EHR) & 0.4875           \\ \hline
\multirow{2}{*}{PECon (Ours)} & Pretraining & 100    & \multirow{2}{*}{SGD} & 0.1                                                      & -                \\
                              & Finetuning  & 25     &                      & 0.01(CT), 0.1(EHR) & 0.375            \\ \hline
\end{tabular}
\label{tab:supplementary}
\end{table}

\begin{figure}[h]
\centering
\includegraphics[width=0.8\textwidth]{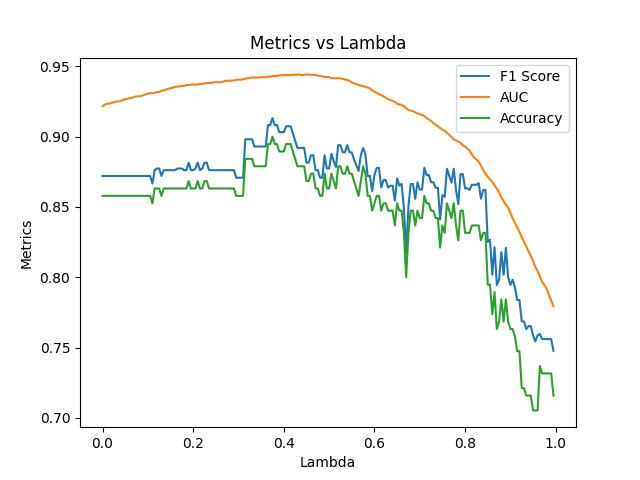}
\caption{Ablation study on the $\lambda$ hyper-parameter}
\label{fig:lambda}
\end{figure}

\begin{table}[]
\centering
\label{tab:ablation-3}
\caption{ The table describes the comparison of Cross-Entropy loss ($\gamma = 0 $) with Focal Loss with a varying $\gamma$ value. CE represents the cross entropy loss.}
\begin{tabular}{cccc}
\hline
Loss ($\gamma$) & Accuracy & F1 score & AUROC \\ \hline
\textbf{0 (CE)}         &   \textbf{0.900}       &   \textbf{0.913}       &   \textbf{0.943}    \\
0.5          &   0.889       &    0.902      &   0.941    \\
1            &   0.884       &    0.898      &   0.935    \\
2            &   0.889       &    0.905      &   0.927    \\ \hline
\end{tabular}
\end{table}

\bibliographystyle{splncs05}
% \bibliography{refs}